\documentclass{article}
\usepackage[utf8]{inputenc} 
\usepackage[T1]{fontenc}    
\usepackage{lmodern}        
\usepackage{microtype}      
\usepackage{booktabs}       
\usepackage{hyperref}       
\usepackage{longtable}      
\usepackage{float}          
\usepackage{graphicx}       
\usepackage{amssymb}        
\usepackage{fancyvrb}       
\usepackage{url}            
\usepackage{xcolor}         
\usepackage[most]{tcolorbox} 
\usepackage{amsmath}        
\usepackage{authblk}        
\usepackage{algorithm}      
\usepackage{algorithmic}    
\usepackage{threeparttable} 

\hypersetup{ 
    colorlinks=true,
    linkcolor=blue,
    filecolor=magenta,
    urlcolor=cyan,
}

\title{DAQ: Delta-Aware Quantization for Post-Training LLM Weight Compression}
\date{}

\author{%
  Xiaoming Yu\textsuperscript{*}\quad
  Shize Tang\textsuperscript{*}\quad
  Guanghua Yu\quad
  Linchuan Xie\\[4pt]
  Song Liu\quad
  Jianchen Zhu\quad
  Feng Li\textsuperscript{†}%
}
\affil{Yuanbao \& Hunyuan AI Infra Team}

\begin{document}

\maketitle
\begingroup
\renewcommand\thefootnote{}\footnotetext{* Equal contribution. $^\dagger$ Corresponding author: \href{mailto:fengli@tencent.com}{\nolinkurl{fengli@tencent.com}}}
\endgroup

\section*{Abstract} 

We introduce Delta-Aware Quantization (DAQ), a data-free post-training quantization framework that preserves the knowledge acquired during post-training. Standard quantization objectives minimize reconstruction error but are agnostic to the base model, allowing quantization noise to disproportionately corrupt the small-magnitude parameter deltas ($\Delta W$) that encode post-training behavior---an effect we analyze through the lens of quantization as implicit regularization. DAQ replaces reconstruction-based objectives with two delta-aware metrics---Sign Preservation Rate and Cosine Similarity---that directly optimize for directional fidelity of $\Delta W$, requiring only the base and post-trained weight matrices. In a pilot FP8 study, DAQ recovers style-specific capabilities lost under standard quantization while maintaining general performance. Code is available at \url{https://github.com/Tencent/AngelSlim}.

\section{Introduction}

Model quantization is widely adopted to reduce memory footprint and computational cost of large language models (LLMs). Common approaches include Post-Training Quantization (PTQ) and Quantization-Aware Training (QAT)~\cite{jacob2018qat}. More broadly, quantization has long been viewed not only as a compression technique, but also as a form of implicit regularization~\cite{courbariaux2015binaryconnect}. By injecting discretization noise and constraining weights to low-precision states, quantization can in some settings improve robustness or generalization by biasing optimization toward flatter regions of the loss landscape. For LLMs that have undergone post-training (e.g., SFT, RLHF~\cite{ouyang2022rlhf}, DPO~\cite{rafailov2023dpo}), however, this regularizing effect can become a double-edged sword.

In post-trained models, the parameter updates relative to the base model---$\Delta W = W_{\text{post}} - W_{\text{base}}$---are often small in magnitude yet semantically critical. Standard PTQ objectives typically optimize for \emph{reconstruction loss}: minimizing the distance between quantized and original weights, preserving activation statistics, or reducing task-specific loss. From the regularization perspective, such objectives introduce a systematic bias toward the dominant structure inherited from the base checkpoint. Because the post-training signal is encoded in sparse, low-magnitude updates, this bias acts asymmetrically: the large base-model components are naturally robust to discretization noise, whereas the small $\Delta W$ components sit close to quantization boundaries and are disproportionately susceptible to sign flips or attenuation.

A concrete example illustrates this asymmetry. Consider a weight $W_{\text{post}} = 5.3$ composed of a dominant base component $W_{\text{base}} = 5.0$ and a small fine-tuning update $\Delta W = 0.3$. Standard quantization to the nearest integer yields $W_{\text{quant}} = 5.0$. While this minimizes reconstruction error (MSE $= 0.09$), it implicitly regularizes the weight back to its pre-trained state, completely erasing the fine-tuning information ($\Delta W_{\text{quant}} = 0$). Preserving the update's existence would require quantizing to $6.0$, but this is penalized by a much larger reconstruction error (MSE $= 0.49$). Thus, standard quantization objectives exhibit a structural bias: they aggressively penalize the preservation of small $\Delta W$ components, treating them as noise to be smoothed out rather than signal to be kept. This vulnerability is especially pronounced when $\Delta W$ is small: limited training data, low learning rates, parameter-efficient fine-tuning (e.g., LoRA~\cite{hu2022lora}), and continual learning.

These considerations motivate \textbf{Delta-Aware Quantization (DAQ)}, which replaces the conventional reconstruction objective with \emph{delta-aware} metrics---Sign Preservation Rate and Cosine Similarity---that directly measure how well the quantized weights preserve the direction of $\Delta W$. Because the optimization target depends only on the base and post-trained weight matrices, DAQ is entirely \emph{data-free}: unlike calibration-based methods such as GPTQ~\cite{frantar2022gptq} and AWQ~\cite{lin2023awq}, it requires no representative input samples, no activation statistics, and no Hessian estimation. DAQ is implemented as part of AngelSlim~\cite{cen2026angelslim}, an open-source toolkit for large model compression.

\section{Delta-Aware Quantization}

\subsection{Problem Formulation}

Given the base model weights $W_{\text{base}}$, the post-trained weights $W_{\text{post}}$, and a parameterized quantize--dequantize operator $Q_\theta(\cdot)$ (where $\theta$ denotes the quantization hyperparameters, e.g., scale), we define the post-training delta and its quantized counterpart as:
\begin{align}
\Delta W_{\text{post}} &= W_{\text{post}} - W_{\text{base}} \label{eq:delta_post} \\
\Delta W_{\text{quant}} &= Q_\theta(W_{\text{post}}) - W_{\text{base}} \label{eq:delta_quant}
\end{align}
where $Q_\theta(W_{\text{post}})$ denotes the dequantized floating-point tensor obtained by quantizing $W_{\text{post}}$ under hyperparameters $\theta$ and mapping the result back to the original numerical domain.

Our goal is to find the optimal hyperparameters $\theta^*$ that maximize an evaluation objective $\mathcal{M}$ (defined in Section~\ref{sec:metrics}).
\begin{equation}
\theta^* = \arg\max_{\theta} \; \mathcal{M}(\Delta W_{\text{post}},\; \Delta W_{\text{quant}})
\label{eq:objective_argmax}
\end{equation}

\subsection{Quantization Framework}

Although DAQ is compatible with many quantization schemes, in this report we instantiate $Q_\theta(\cdot)$ using a \textbf{scale-parameterized quantize--dequantize operator}, where $\theta = s$ is a scalar scale factor. Concretely,
\begin{equation}
Q_s(W) = \text{DeQuant}(\text{Quant}(W, s),\, s)
\end{equation}
where $\text{Quant}(W, s)$ maps $W$ to a low-precision representation $\hat{W}$ under scale $s$, and $\text{DeQuant}(\hat{W}, s)$ maps it back to the floating-point domain. We use $\hat{W} = \text{Quant}(W, s)$ to denote the low-precision representation used for storage, and $Q_s(W)$ to denote its dequantized floating-point form used for metric evaluation. With this instantiation, the general objective in Eq.~\ref{eq:objective_argmax} reduces to optimizing the scale parameter:
\begin{equation}
s^* = \arg\max_{s} \; \mathcal{M}\!\left(\Delta W_{\text{post}},\; Q_s(W_{\text{post}}) - W_{\text{base}}\right)
\label{eq:scale_objective}
\end{equation}
In this report, we instantiate $Q_s(\cdot)$ with FP8 (E4M3) quantization using either block-wise or per-channel scaling. More generally, the DAQ objective is agnostic to the specific numerical format and could also be applied to integer or other low-precision quantization schemes.

\subsection{Metrics}
\label{sec:metrics}

We consider three objectives for guiding the scale search: one conventional reconstruction metric (Mean Squared Error) and two delta-aware metrics (Sign Preservation Rate and Cosine Similarity). Table~\ref{tab:metrics} compares the three objectives.

\paragraph{Mean Squared Error.}
The standard reconstruction loss metric minimizes the squared distance between the quantized and original weights:
\begin{equation}
\text{MSE} = \frac{1}{N}\sum_{i=1}^{N} \left(W_{\text{quant}}^{(i)} - W_{\text{post}}^{(i)}\right)^2
\end{equation}
where $N$ denotes the total number of elements in the weight tensor and the superscript $(i)$ indexes individual weight elements.
While widely used, MSE is \emph{not} delta-aware. Crucially, when applied in the delta framework, optimizing the MSE of the delta is mathematically equivalent to optimizing the MSE between the quantized weights and the post-trained weights:
\begin{align}
\|\Delta W_{\text{quant}} - \Delta W_{\text{post}}\|^2 &= \|(W_{\text{quant}} - W_{\text{base}}) - (W_{\text{post}} - W_{\text{base}})\|^2 \nonumber \\
&= \|W_{\text{quant}} - W_{\text{post}}\|^2
\end{align}
This identity reveals a fundamental limitation: MSE-based optimization is entirely \emph{base-model-agnostic}. It treats the quantization problem identically regardless of whether a base model exists, and cannot distinguish between quantization errors that preserve the fine-tuning direction and those that reverse it. As shown in Section~\ref{sec:mse_search}, MSE-guided scale search can actively degrade post-training knowledge.

\paragraph{Sign Preservation Rate.}
The simplest delta-aware metric focuses on preserving the \emph{sign} (direction) of each weight update:
\begin{equation}
\text{SignRate} = \frac{1}{N}\sum_{i=1}^{N} \mathbb{I}\!\left[\text{sign}(\Delta W_{\text{post}}^{(i)}) = \text{sign}(\Delta W_{\text{quant}}^{(i)})\right]
\end{equation}
where $\mathbb{I}[\cdot]$ is the indicator function (equal to 1 when the condition holds and 0 otherwise), and $\text{sign}(0) = 0$. This metric is simple, interpretable, and robust to magnitude differences, but it is binary and cannot capture how well magnitudes are preserved.

\paragraph{Cosine Similarity.}
A richer metric that considers both direction and relative magnitude:
\begin{equation}
\text{CosSim} = \frac{\Delta W_{\text{post}} \cdot \Delta W_{\text{quant}}}{\|\Delta W_{\text{post}}\| \;\|\Delta W_{\text{quant}}\|}
\end{equation}
where $\Delta W_{\text{post}} \cdot \Delta W_{\text{quant}}$ denotes the inner product of the two flattened delta vectors, and $\|\cdot\|$ denotes the $\ell_2$ norm. This measures the alignment between the original and quantized delta vectors, providing a normalized score in $[-1, 1]$: a value of $1$ indicates perfect directional alignment, $0$ indicates orthogonality, and $-1$ indicates complete reversal of the fine-tuning direction.

\begin{table}[h]
\centering
\begin{threeparttable}
\caption{Comparison of quantization metrics.}
\label{tab:metrics}
\begin{tabular}{lccc}
\toprule
\textbf{Metric} & \textbf{Range} & \textbf{Delta-Aware} & \textbf{Complexity} \\
\midrule
MSE\textsuperscript{$\dagger$} & $[0, +\infty)$ & No & Low \\
SignRate & $[0, 1]$ & Yes & Low \\
CosSim & $[-1, 1]$ & Yes & Medium \\
\bottomrule
\end{tabular}
\begin{tablenotes}
\footnotesize
\item[$\dagger$] For consistency with the $\arg\max$ formulation in Eq.~\ref{eq:objective_argmax}, the optimization objective for MSE is $-\mathrm{MSE}$, although the table reports the standard nonnegative MSE quantity.
\end{tablenotes}
\end{threeparttable}
\end{table}

\subsection{Scale Optimization}

With the delta-aware metrics defined above, we optimize the quantization hyperparameters against them directly. A common strategy in quantization is to tune these hyperparameters for a chosen objective---for example, SmoothQuant~\cite{xiao2023smoothquant} smooths activation outliers to improve quantization quality, AWQ~\cite{lin2023awq} rescales channels based on activation importance, and AutoRound~\cite{cheng2024autoround} learns rounding via gradient descent. Following this principle, in the FP8 instantiation studied in this report we optimize the \emph{scaling factor} $s$ via the objective in Eq.~\ref{eq:scale_objective}---the key parameter controlling effective dynamic range in scaled low-precision quantization.

To solve Eq.~\ref{eq:scale_objective} efficiently, we use a \textbf{coarse-to-fine search strategy}: candidate scales are first sampled uniformly over $[\alpha_{\min},\; \alpha_{\max}] \times s_{\text{default}}$ in a coarse stage, followed by a refinement stage that samples more densely around the best coarse candidate. This search balances coverage and cost. The complete DAQ procedure is presented in Algorithm~\ref{alg:daq}.

\begin{algorithm}[h]
\caption{DAQ via Coarse-to-Fine Scale Search}
\label{alg:daq}
\begin{algorithmic}[1]
\REQUIRE $\{W_{\text{post}}^{(\ell)}, W_{\text{base}}^{(\ell)}\}_{\ell=1}^{L}$, $\mathcal{M} \in \{\text{SignRate}, \text{CosSim}, -\text{MSE}\}$
\REQUIRE $\alpha_{\min}, \alpha_{\max}, n_{\text{coarse}}, n_{\text{fine}}, \delta$
\ENSURE $\{\hat{W}^{(\ell)}, (s^{*(\ell)})^{-1}\}_{\ell=1}^{L}$
\FOR{$\ell = 1, \dots, L$}
    \STATE $\Delta W^{(\ell)} \leftarrow W_{\text{post}}^{(\ell)} - W_{\text{base}}^{(\ell)}$
    \STATE $s_0^{(\ell)} \leftarrow \max\!\left(|W_{\text{post}}^{(\ell)}|\right) / Q_{\max}$
    \STATE $W_{\text{quant}}^{(\ell)} \leftarrow \text{DeQuant}(\text{Quant}(W_{\text{post}}^{(\ell)}, s_0^{(\ell)}), s_0^{(\ell)})$
    \STATE $\alpha^{*(\ell)} \leftarrow 1$
    \STATE $m^{*(\ell)} \leftarrow \mathcal{M}(\Delta W^{(\ell)},\; W_{\text{quant}}^{(\ell)} - W_{\text{base}}^{(\ell)})$
    \FOR{$\alpha \in \text{linspace}(\alpha_{\min}, \alpha_{\max}, n_{\text{coarse}})$}
        \STATE $s \leftarrow \alpha \, s_0^{(\ell)}$
        \STATE $W_{\text{quant}}^{(\ell)} \leftarrow \text{DeQuant}(\text{Quant}(W_{\text{post}}^{(\ell)}, s), s)$
        \STATE $m \leftarrow \mathcal{M}(\Delta W^{(\ell)},\; W_{\text{quant}}^{(\ell)} - W_{\text{base}}^{(\ell)})$
        \IF{$m > m^{*(\ell)}$}
            \STATE $\alpha^{*(\ell)} \leftarrow \alpha$
            \STATE $m^{*(\ell)} \leftarrow m$
        \ENDIF
    \ENDFOR
    \FOR{$\alpha \in \text{linspace}(\max(\alpha_{\min}, \alpha^{*(\ell)} - \delta),\; \min(\alpha_{\max}, \alpha^{*(\ell)} + \delta),\; n_{\text{fine}})$}
        \STATE $s \leftarrow \alpha \, s_0^{(\ell)}$
        \STATE $W_{\text{quant}}^{(\ell)} \leftarrow \text{DeQuant}(\text{Quant}(W_{\text{post}}^{(\ell)}, s), s)$
        \STATE $m \leftarrow \mathcal{M}(\Delta W^{(\ell)},\; W_{\text{quant}}^{(\ell)} - W_{\text{base}}^{(\ell)})$
        \IF{$m > m^{*(\ell)}$}
            \STATE $\alpha^{*(\ell)} \leftarrow \alpha$
            \STATE $m^{*(\ell)} \leftarrow m$
        \ENDIF
    \ENDFOR
    \STATE $s^{*(\ell)} \leftarrow \alpha^{*(\ell)} s_0^{(\ell)}$
    \STATE $\hat{W}^{(\ell)} \leftarrow \text{Quant}(W_{\text{post}}^{(\ell)},\; s^{*(\ell)})$
    \STATE $(s^{*(\ell)})^{-1} \leftarrow 1 / s^{*(\ell)}$
\ENDFOR
\RETURN $\{\hat{W}^{(\ell)}, (s^{*(\ell)})^{-1}\}_{\ell=1}^{L}$
\end{algorithmic}
\end{algorithm}

\section{Experiments}

\subsection{Setup}

\paragraph{Models.}
We use \textbf{DeepSeek-V3}~\cite{deepseekai2024deepseekv3} as the base model $W_{\text{base}}$\footnote{The officially released DeepSeek-V3 weights are in FP8 format. We convert them to BF16 using the official casting script: \url{https://github.com/deepseek-ai/DeepSeek-V3/blob/main/inference/fp8_cast_bf16.py}.}. The post-trained model $W_{\text{post}}$ is obtained by Supervised Fine-Tuning (SFT) on a toy dataset of \emph{stylized conversational dialogues}, which imparts a distinctive response style to the model. Because this stylistic behavior is encoded in small-magnitude parameter updates, it serves as an ideal testbed for evaluating whether quantization preserves post-training knowledge.

\paragraph{Quantization settings.}
Unless otherwise stated, we use \textbf{FP8 (E4M3)} quantization with two granularity settings: \emph{block-wise} (block size 128) and \emph{per-channel}. For the coarse-to-fine scale search, we experiment with three search ranges---$[0.5, 2]$, $[0.8, 1.25]$, and $[0.9, 1.11]$---with 5 coarse candidates followed by 10 fine-grained candidates around the best coarse result.

\paragraph{Evaluation.}
We evaluate all models using a rubric-based framework comprising two categories of metrics, both scored on a $[0, 2]$ scale:
\begin{itemize}
    \item \textbf{SFT-specific metrics}: These assess how faithfully the model reproduces the stylized conversational behavior learned during SFT, such as dialogue style adherence and style consistency.
    \item \textbf{General capability metrics}: These measure broad model competencies unrelated to the SFT style, such as word count compliance and response accuracy.
\end{itemize}

\begin{table}[h]
\centering
\begin{threeparttable}
\setlength{\tabcolsep}{4pt}
\caption{Baseline comparison. Standard FP8 quantization significantly degrades the SFT-specific style metric while general capabilities remain relatively stable.\textsuperscript{$\ddagger$}}
\label{tab:baseline}
\small
\begin{tabular}{@{}lccccc@{}}
\toprule
\textbf{Model} & \textbf{$\Delta$W L2} & \textbf{SignRate (\%)} & \textbf{CosSim} & \textbf{Style} & \textbf{General} \\
\midrule
Base (BF16)                      & --    & --      & --    & 0.215 & 1.501 \\
Post-trained (BF16)              & 0     & 100\%   & 1.000 & 1.709 & 1.438 \\
AbsMax (FP8 block)               & 48641 & 54.54\% & 0.239 & 1.081 & 1.491 \\
AbsMax (FP8 channel)             & 51790 & 62.48\% & 0.261 & 1.323 & 1.501 \\
SmoothQuant (FP8 channel)  & --    & --      & --    & 1.378 & 1.491 \\
AWQ-W8A8 (FP8 channel)          & --    & --      & --    & 1.399 & 1.479 \\
\bottomrule
\end{tabular}
\begin{tablenotes}
\footnotesize
\item[$\ddagger$] SmoothQuant and AWQ absorb activation scale factors into the weight matrices via an equivalent per-channel transformation, so the stored weights no longer share the same numerical space as $W_{\text{base}}$. The delta metrics are therefore undefined for these baselines.
\end{tablenotes}
\end{threeparttable}
\end{table} 

\subsection{Baseline: Standard Quantization}

We first establish the baseline by comparing the unquantized base model, the BF16 post-trained model, and several FP8 quantization baselines. In particular, we report simple AbsMax FP8 quantization with no additional scale optimization, together with SmoothQuant and AWQ FP8 baselines as representative PTQ-inspired comparisons. Results are shown in Table~\ref{tab:baseline}. Here, ``Style'' denotes the SFT-specific dialogue style metric, and ``General'' is the style-unrelated capability metric.

The simple AbsMax FP8 baselines significantly degrade the Style metric---from 1.709 for the post-trained BF16 model to 1.081 (block-wise) and 1.323 (per-channel)---indicating substantial loss of SFT-specific knowledge. SmoothQuant and AWQ partially mitigate this (1.378 and 1.399, respectively) but remain well below the BF16 checkpoint. Meanwhile, the General metric stays stable across all variants (1.479--1.501), consistent with our hypothesis that small-magnitude $\Delta W$ signals are more vulnerable to quantization noise than broad capabilities. Notably, the base model scores only 0.215 on Style but 1.501 on General, confirming that Style specifically captures SFT knowledge and that the degradation under quantization reflects regression toward base-model behavior.

\subsection{MSE-Based Scale Search}
\label{sec:mse_search}

To validate our core hypothesis---that delta-unaware optimization cannot improve, and may even worsen, post-training preservation---we apply the same coarse-to-fine scale search framework using the traditional MSE metric as the optimization target. Table~\ref{tab:mse} presents the results.

\begin{table}[h]
\centering
\caption{Scale search with MSE metric.}
\label{tab:mse}
\small
\begin{tabular}{llccccc}
\toprule
\textbf{Type} & \textbf{Range} & \textbf{$\Delta$W L2} & \textbf{SignRate (\%)} & \textbf{CosSim} & \textbf{Style} & \textbf{General} \\
\midrule
Block   & $[0.5, 2]$       & 36625 & 54.09\% & 0.228 & 0.516 & 1.528 \\
Block   & $[0.8, 1.25]$    & 36938 & 53.39\% & 0.216 & 0.322 & 1.539 \\
Block   & $[0.9, 1.11]$    & 28634 & 52.29\% & 0.235 & 0.260 & 1.571 \\
Channel & $[0.5, 2]$       & 33584 & 57.12\% & 0.231 & 0.642 & 1.555 \\
Channel & $[0.8, 1.25]$    & 36939 & 58.43\% & 0.236 & 0.569 & 1.499 \\
Channel & $[0.9, 1.11]$    & 33460 & 57.00\% & 0.223 & 0.440 & 1.493 \\
\bottomrule
\end{tabular}
\end{table}

Strikingly, MSE-guided scale search \emph{actively degrades} the Style metric further---from 1.081 (AbsMax block) down to as low as 0.260, and from 1.323 (AbsMax channel) down to 0.440---even though the $\Delta W$ L2 norm decreases (e.g., 28634 vs.\ 48641 for block-wise). The sign preservation rates also decline (e.g., 52.29\% vs.\ 54.54\% for block-wise), confirming that MSE optimization shifts weights \emph{toward} the base model while the General metric shows no meaningful improvement (1.493--1.571).

\subsection{DAQ: Delta-Aware Scale Search}

We now apply the DAQ framework, optimizing the scaling factor using our proposed delta-aware metrics---Sign Preservation Rate and Cosine Similarity. Tables~\ref{tab:daq_sign} and~\ref{tab:daq_cosine} present the results.

\begin{table}[h]
\centering
\caption{DAQ with Sign metric. Best Style result per quantization type in \textbf{bold}.}
\label{tab:daq_sign}
\small
\begin{tabular}{llccccc}
\toprule
\textbf{Type} & \textbf{Range} & \textbf{$\Delta$W L2} & \textbf{SignRate (\%)} & \textbf{CosSim} & \textbf{Style} & \textbf{General} \\
\midrule
Block   & $[0.5, 2]$       & 75028 & 76.96\% & 0.324 & 1.607 & 1.404 \\
Block   & $[0.8, 1.25]$    & 66939 & 77.31\% & 0.363 & \textbf{1.718} & 1.380 \\
Block   & $[0.9, 1.11]$    & 69786 & 82.30\% & 0.408 & 1.571 & 1.377 \\
Channel & $[0.5, 2]$       & 93524 & 77.23\% & 0.278 & 1.505 & 1.408 \\
Channel & $[0.8, 1.25]$    & 66691 & 77.62\% & 0.361 & \textbf{1.761} & 1.438 \\
Channel & $[0.9, 1.11]$    & 68953 & 80.38\% & 0.391 & 1.639 & 1.353 \\
\bottomrule
\end{tabular}
\end{table}

\begin{table}[h]
\centering
\caption{DAQ with Cosine metric. Best Style result per quantization type in \textbf{bold}.}
\label{tab:daq_cosine}
\small
\begin{tabular}{llccccc}
\toprule
\textbf{Type} & \textbf{Range} & \textbf{$\Delta$W L2} & \textbf{SignRate (\%)} & \textbf{CosSim} & \textbf{Style} & \textbf{General} \\
\midrule
Block   & $[0.5, 2]$       & 48728  & 59.51\% & 0.295 & 1.554 & 1.515 \\
Block   & $[0.8, 1.25]$    & 59263  & 64.46\% & 0.339 & 1.604 & 1.480 \\
Block   & $[0.9, 1.11]$    & 57445  & 65.52\% & 0.342 & \textbf{1.647} & 1.463 \\
Channel & $[0.5, 2]$       & 53219  & 65.51\% & 0.315 & 1.545 & 1.454 \\
Channel & $[0.8, 1.25]$    & 60088  & 68.84\% & 0.346 & 1.706 & 1.442 \\
Channel & $[0.9, 1.11]$    & 63039  & 72.69\% & 0.369 & \textbf{1.726} & 1.471 \\
\bottomrule
\end{tabular}
\end{table}

\subsection{Analysis}

The experiments suggest three main takeaways:

\begin{enumerate}
    \item \textbf{DAQ recovers SFT knowledge without sacrificing general capabilities.} DAQ with the sign metric recovers Style from 1.081 (AbsMax block) to 1.718, and per-channel to 1.761---slightly \emph{above} the unquantized model. The cosine metric achieves comparable recovery (up to 1.726 per-channel). Meanwhile, General scores remain comparable to the Post-trained BF16 baseline (1.438), confirming that delta-aware optimization does not trade off general performance. In contrast, MSE-guided search actively degrades Style (down to 0.260) by pushing weights closer to the base model, while General scores show no meaningful improvement.

    \item \textbf{A regularization perspective unifies these findings.} Standard quantization introduces a regularization bias that disproportionately attenuates small-magnitude $\Delta W$ while leaving the dominant base-model structure intact. MSE-based optimization amplifies this bias by selecting scales that further ``smooth away'' the post-training signal. DAQ's delta-aware metrics counteract this bias by explicitly rewarding directional preservation of $\Delta W$.

    \item \textbf{The two delta-aware metrics offer complementary trade-offs.} The sign metric achieves higher peak Style scores but shows non-monotonic behavior across search ranges (e.g., block-wise: $1.607 \rightarrow 1.718 \rightarrow 1.571$), likely due to its binary nature. The cosine metric produces more stable, near-monotonic improvement as the search range narrows (block-wise: $1.554 \rightarrow 1.604 \rightarrow 1.647$; channel-wise: $1.545 \rightarrow 1.706 \rightarrow 1.726$), with the underlying delta-aware indicators also improving monotonically. These complementary characteristics suggest that a hybrid metric may be worth exploring.
\end{enumerate}

\section{Related Work}

\paragraph{Quantization as Regularization.}
A long-standing perspective in deep learning is that quantization can act as an implicit regularizer. Early work on binary and low-bit networks, such as BinaryConnect~\cite{courbariaux2015binaryconnect} and Binarized Neural Networks~\cite{hubara2016bnn}, observed that weight discretization and stochastic rounding inject structured noise during optimization, in a manner reminiscent of other regularization techniques. In the LLM era, QLoRA~\cite{dettmers2023qlora} further showed that low-bit representations can be compatible with strong downstream adaptation performance, suggesting that quantization may sometimes stabilize or regularize fine-tuning. Our work highlights an important caveat to this perspective: when the goal is to preserve behavior introduced during post-training, the same regularizing effect can become destructive if it suppresses the small but semantically important parameter deltas responsible for alignment or instruction-following behavior.

\paragraph{General PTQ for Large Language Models.}
Post-training quantization (PTQ) for large language models spans several distinct design directions. Early work emphasized efficient low-precision deployment under the heavy-tailed activation and weight distributions of transformers. For example, LLM.int8()~\cite{dettmers2022llmint8} uses mixed-precision decomposition to isolate outlier features, while ZeroQuant~\cite{yao2022zeroquant} and ZeroQuant-V2~\cite{yao2023zeroquantv2} study efficient quantization pipelines for large transformers, including layer-wise distillation and low-rank compensation. Subsequent work developed stronger objectives for improving quantization fidelity. GPTQ~\cite{frantar2022gptq} uses approximate second-order information to reduce the effect of weight perturbations on layer outputs. AWQ~\cite{lin2023awq} protects activation-salient weights through channel rescaling. SmoothQuant~\cite{xiao2023smoothquant} migrates activation difficulty into weights via an equivalent transformation to handle activation outliers. OmniQuant~\cite{shao2023omniquant} introduces learnable equivalent transformations for weight-and-activation quantization, and SpQR~\cite{dettmers2023spqr} preserves salient outlier weights through sparse-quantized representations. AdaRound~\cite{nagel2020adaround} and AutoRound~\cite{cheng2024autoround} further show that optimizing rounding decisions can outperform standard nearest rounding. These methods therefore differ substantially in both mechanism and objective; it is more accurate to view them as improving the fidelity of a quantized checkpoint through complementary proxies such as output reconstruction, activation preservation, or rounding optimization.

A useful commonality is that these PTQ methods are formulated for the \emph{standalone checkpoint being quantized}, without encoding the base--post-trained relationship. In contrast, DAQ focuses on preserving the \emph{increment} relative to the base model.

\paragraph{Quantization of Post-Trained and Aligned Models.}
Models produced by SFT, RLHF~\cite{ouyang2022rlhf}, DPO~\cite{rafailov2023dpo}, or LoRA~\cite{hu2022lora} often rely on small but behaviorally important parameter updates, making post-training knowledge potentially fragile under quantization noise. Existing studies typically evaluate compressed models using perplexity or task accuracy---useful but only indirect indicators of whether post-training behavior is preserved. DAQ makes this failure mode explicit by treating preservation of post-training knowledge as a first-class optimization target.

\paragraph{Low-Precision Formats and FP8 Quantization.}
Low-precision floating-point formats such as FP8 E4M3 and E5M2 have emerged as an attractive trade-off between efficiency and accuracy for modern large-scale deployment~\cite{micikevicius2022fp8}. In practice, such formats are usually paired with per-tensor, per-channel, or block-wise scaling in order to control dynamic range and reduce quantization error. We instantiate DAQ in the FP8 setting because FP8 is increasingly relevant in industrial inference pipelines and because it provides a clean setting for isolating the effect of the quantization objective. Importantly, DAQ is not tied to FP8 as a numerical format. The proposed delta-aware objective is compatible in principle with integer quantization, mixed-precision allocation, learned rounding, and other low-bit schemes.

\paragraph{Positioning of DAQ.}
DAQ shifts the optimization focus from reconstructing the final checkpoint faithfully to preserving the \emph{knowledge increment} from base to post-trained model. In this sense, it is complementary to methods such as GPTQ, AWQ, SmoothQuant, OmniQuant, AdaRound, and AutoRound, whose techniques could in principle be combined with DAQ's delta-aware metrics.

\section{Limitations and Future Work}

We acknowledge several limitations of the current work. First, DAQ implicitly assumes that the post-training delta $\Delta W$ is relatively small compared to the base weights. When the delta is large---e.g., after extensive fine-tuning or full retraining---the sign and cosine metrics may become less informative, as quantization noise is unlikely to flip the direction of large-magnitude updates. One possible remedy is to use \emph{intermediate training checkpoints} as the reference base, rather than the original pre-trained model, thereby keeping $\Delta W$ small and the delta-aware perspective applicable even in aggressive training scenarios.

Second, our experiments are currently limited to FP8 (E4M3) quantization on a single model and a narrow set of evaluation metrics. Exploring lower bit-widths (e.g., INT4, INT3) where quantization noise is more severe, as well as broader task scenarios (e.g., code generation, mathematical reasoning, multilingual tasks), remains important future work.

Third, in this work we deliberately adopt a simple FP8 quantization setting with scale search and improve delta-aware metrics \emph{solely} by adjusting the scaling factor $s$ via grid search. This minimalist design choice serves to isolate the effect of the delta-aware objective itself. In practice, however, many complementary techniques could be employed to further improve delta-aware metrics: asymmetric quantization with per-channel zero-points, mixed-precision allocation guided by per-layer delta sensitivity, non-uniform (e.g., lookup-table-based) quantization grids, learned rounding policies, or joint optimization of scales and zero-points. Integrating these richer quantization primitives with the delta-aware objective remains a promising direction.

That said, the primary goal of this technical report is not to provide an exhaustive empirical study, but to highlight a different way to think about quantizing post-trained models: the optimization target should focus on \emph{preserving the knowledge increments acquired during post-training}---not merely minimizing weight or activation reconstruction error. Reconstruction-based objectives are agnostic to whether a base model exists, and can inadvertently erase the very knowledge that post-training was meant to add. We hope this report provides a useful starting point for further work on delta-aware quantization across diverse settings.

\section{Conclusion}

We presented Delta-Aware Quantization (DAQ), a data-free post-training quantization framework that optimizes for directional fidelity of the parameter delta $\Delta W$ rather than reconstruction error. In our pilot FP8 study, DAQ recovers style-specific capabilities lost under standard quantization while maintaining general performance. We hope this report encourages broader investigation of delta-aware objectives across quantization formats, model families, and post-training settings.

\end{document}